\pdfoutput=1

\documentclass[11pt]{article}

\usepackage[]{ACL2023}

\usepackage{times}
\usepackage{latexsym}
\usepackage{amsfonts}
\usepackage{amsmath}
\usepackage{amssymb}
\usepackage{multicol}
\usepackage{multirow}
\usepackage{xspace}
\usepackage{booktabs}
\usepackage{bbding}
\usepackage{array}
\usepackage{threeparttable}
\usepackage{tcolorbox}
\usepackage{enumitem}
\usepackage{tabularx}
\usepackage[linesnumbered,ruled,vlined]{algorithm2e}
\usepackage{xcolor,colortbl}
\usepackage{setspace}

\usepackage{todonotes}

\usepackage{cleveref}
\crefformat{section}{\S#2#1#3}
\crefformat{subsection}{\S#2#1#3}
\crefformat{subsubsection}{\S#2#1#3}
\crefrangeformat{section}{\S\S#3#1#4 to~#5#2#6}
\crefmultiformat{section}{\S\S#2#1#3}{ and~#2#1#3}{, #2#1#3}{ and~#2#1#3}
\crefmultiformat{subsection}{\S\S#2#1#3}{ and~#2#1#3}{, #2#1#3}{ and~#2#1#3}
\Crefformat{figure}{#2Fig.~#1#3}
\Crefmultiformat{figure}{Figs.~#2#1#3}{ and~#2#1#3}{, #2#1#3}{ and~#2#1#3}
\Crefformat{table}{#2Tab.~#1#3}
\Crefmultiformat{table}{Tabs.~#2#1#3}{ and~#2#1#3}{, #2#1#3}{ and~#2#1#3}
\Crefformat{appendix}{Appx.~\S#2#1#3}
\crefmultiformat{appendix}{Appx.~\S#2#1#3}{ and~#2#1#3}{, #2#1#3}{ and~#2#1#3}
\crefformat{algorithm}{Alg.~#2#1#3}
\Crefformat{equation}{Eq.~#2#1#3}

\definecolor{ballblue}{rgb}{0.13, 0.67, 0.8}
\definecolor{azure1}{rgb}{0.0, 0.5, 1.0}
\definecolor{uclablue}{rgb}{0.33, 0.41, 0.58}
\definecolor{ultramarine}{rgb}{0.07, 0.04, 0.56}
\definecolor{yaleblue}{rgb}{0.06, 0.3, 0.57}

\newcommand{\xhdr}[1]{\vspace{0.3em}\noindent{{\bf #1.}}}
\newcommand{\modelname}{\textsc{Mr.Cod}\xspace}

\newcommand{\modelnameexplain}{(\underline{M}ulti-hop evidence \underline{r}etrieval for \underline{C}r\underline{o}ss-\underline{d}ocument relation extraction)\xspace}

\usepackage[T1]{fontenc}

\usepackage[utf8]{inputenc}

\usepackage{microtype}

\usepackage{inconsolata}

\title{Multi-hop Evidence Retrieval for Cross-document Relation Extraction}

\author{
Keming Lu,\textsuperscript{\protect\includegraphics[width=0.25cm]{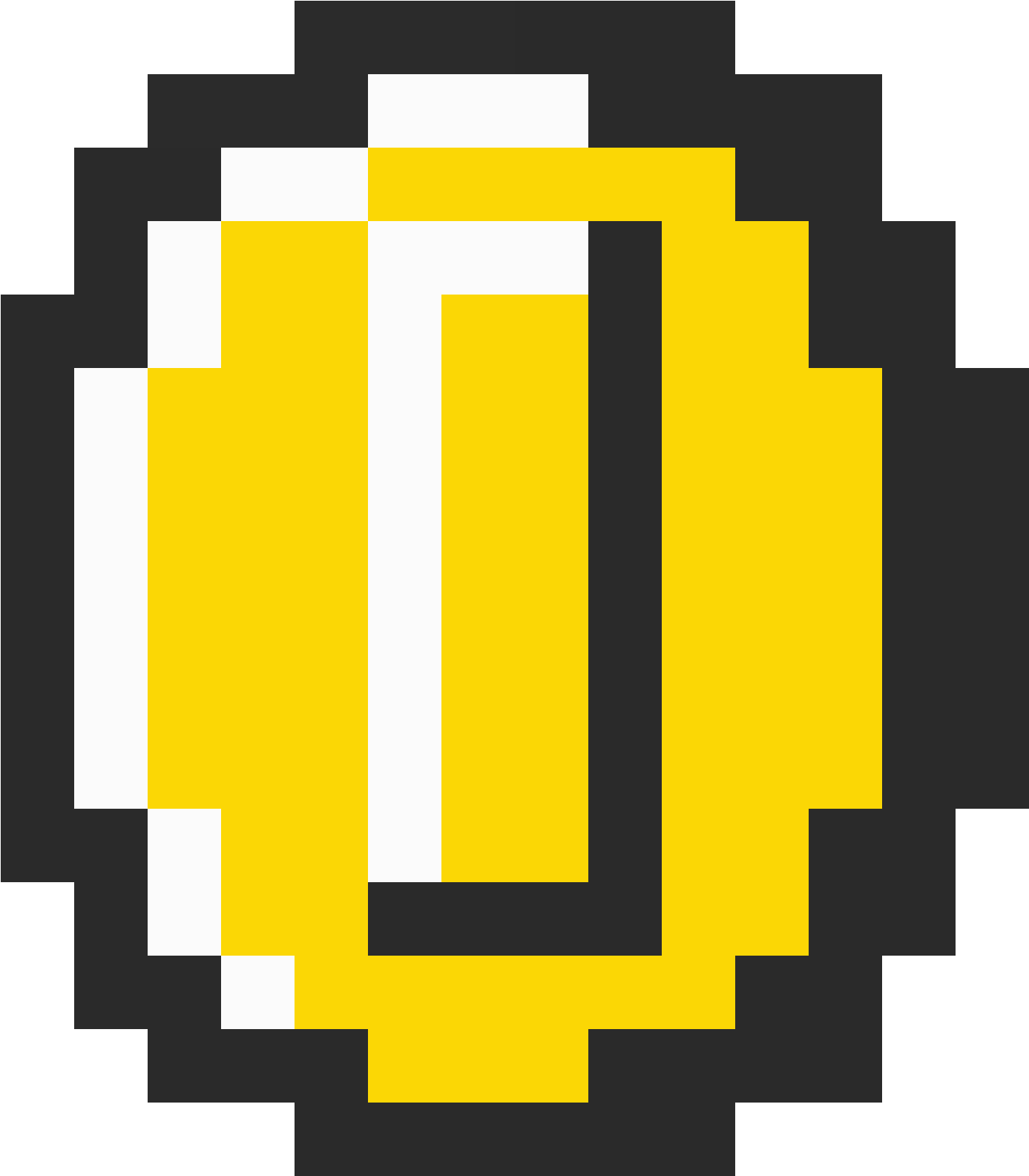}}
I-Hung Hsu,\textsuperscript{\protect\includegraphics[width=0.25cm]{figs/coin_icon.png}}
Wenxuan Zhou,\textsuperscript{\protect\includegraphics[width=0.25cm]{figs/coin_icon.png}}
Mingyu Derek Ma\textsuperscript{\protect\includegraphics[width=0.3cm]{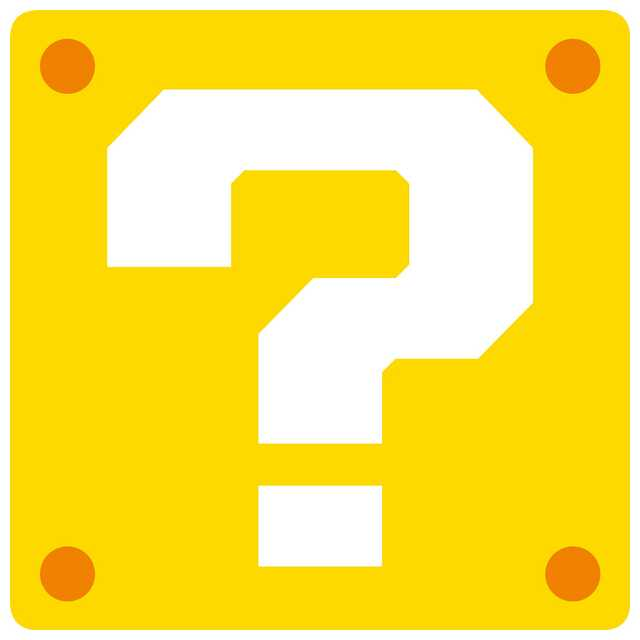}} \and Muhao Chen\textsuperscript{\protect\includegraphics[width=0.25cm]{figs/coin_icon.png}}\\
\textsuperscript{\protect\includegraphics[width=0.25cm]{figs/coin_icon.png}}University of Southern California\\
\textsuperscript{\protect\includegraphics[width=0.3cm]{figs/question_mark_icon.png}}University of California, Los Angeles \\ 
\texttt{\{keminglu,ihunghsu,zhouwenx,muhaoche\}@usc.edu};\; \texttt{{ma@cs.ucla.edu}}\\
}

\begin{document}
\maketitle
\begin{abstract}
Relation Extraction (RE) has been extended to cross-document scenarios because many relations are not simply described in a single document.
This inevitably brings the challenge of efficient open-space evidence retrieval to support the inference of cross-document relations,
along with the challenge of multi-hop reasoning on top of entities and evidence scattered in an open set of documents.
To combat these challenges, we propose \modelname \modelnameexplain, which is a multi-hop evidence retrieval method based on evidence path mining and ranking.
We explore multiple variants of retrievers to show evidence retrieval is essential in cross-document RE.
We also propose a contextual dense retriever for this setting.
Experiments on CodRED show that evidence retrieval with \modelname effectively acquires cross-document evidence and boosts end-to-end RE performance in both closed and open settings.\footnote{Our code is public available at the Github repository: \url{https://github.com/luka-group/MrCoD}}
\end{abstract}

\section{Introduction}
Relation extraction (RE) is a fundamental task of information extraction~\cite{han2020more} that seeks to identify the relation of entities described according to some context.
It is a key task integral to natural language understanding~\cite{liu-etal-2018-knowledge,zhao-etal-2020-knowledge-grounded} for inducing the structural perception of unstructured text.
Furthermore, it is also an essential step of automated knowledge base construction~\cite{niu2012deepdive,subasic2019building} and is the backbone of nearly all knowledge-driven AI tasks~\cite{yasunaga-etal-2021-qa,hao-etal-2017-end,lin-etal-2019-kagnet,DBLP:conf/acl/FungTRPJCMBS20,peters-etal-2019-knowledge}.

\begin{figure}[t]
    \centering
    \includegraphics[width=\linewidth]{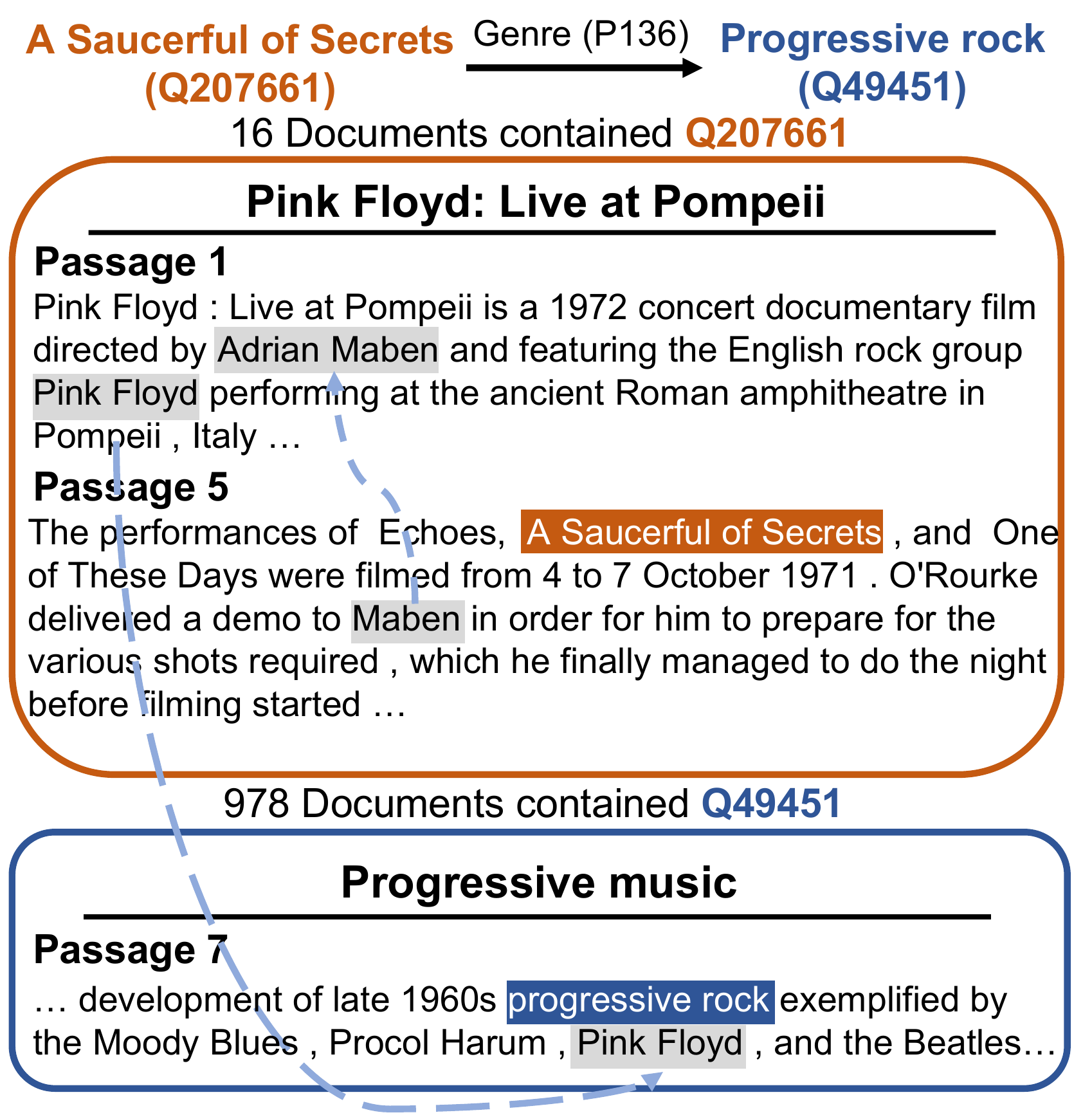}
    \caption{A case for cross-document multi-hop relation reasoning in CodRED.
    This figure shows a 3-hop evidence path for the triplet (``A Saucerful of Secrets'', ``Genre'', ``Progressive rock''), which consists of three passages scattered across two documents from Wikipedia.
    Arrows in the figure show the bridging entities that link the passages in this evidence path.
    }
    \label{fig:data_case}
\end{figure}

Previous works have limited the context of RE within a single sentence~\cite{zhang-etal-2017-position,hsu2022simple,zhou2021improved}, a bag of sentences~\cite{zeng-etal-2015-distant,hsu2021discourse,zhu-etal-2020-nyth}, or a single document~\cite{yao-etal-2019-docred,zhou2021document,tan-etal-2022-document}.
However, more relations can only be established if multiple documents are considered.
For example, more than 57.6\% of the relation facts in Wikidata~\cite{erxleben2014introducing} are not described in individual Wikipedia documents~\cite{yao-etal-2021-codred}.
In addition, humans also consolidate different steps of a complex event by referring to multiple articles, such as inferring the process of an event from multiple news articles~\cite{naughton2006event} or instructional events~\cite{zhang-etal-2020-analogous}.
To facilitate research in cross-document RE, \citet{yao-etal-2021-codred} constructed the first human-annotated cross-document RE dataset, CodRED, to serve as the starting point of this realistic problem.

Unlike sentence- or document-level RE tasks, cross-document RE takes a large corpus of documents as input and poses unique challenges~\cite{yao-etal-2021-codred}.
First, because inferring relations based on the whole corpus is inefficient and impractical, evidence retrieval, which involves extracting evidential context from a large corpus, is crucial for cross-document RE.
Second, relations in cross-document RE are usually described by multi-hop reasoning chains with bridging entities.
\Cref{fig:data_case} shows an example of a 3-hop evidence path in CodRED, which spans three related passages in two documents.
In this example, the relation between `A Saucerful of Secrets'' and ``progressive rock'' is described by a reasoning chain containing four bridging entities (marked in grey).
On average, cross-document multi-hop reasoning through 4.7 bridging entities is needed in CodRED to infer relations~\cite{yao-etal-2021-codred}.
Besides, previous work~\cite{zeng-etal-2020-double,xu-etal-2021-discriminative} has shown that intra-document multi-hop reasoning is effective for document-level RE.
Therefore, evidence retrieval needs to consider the bridging entities for multi-hop reasoning in cross-document RE.

\begin{figure*}[thbp]
    \centering
    \includegraphics[width=\linewidth]{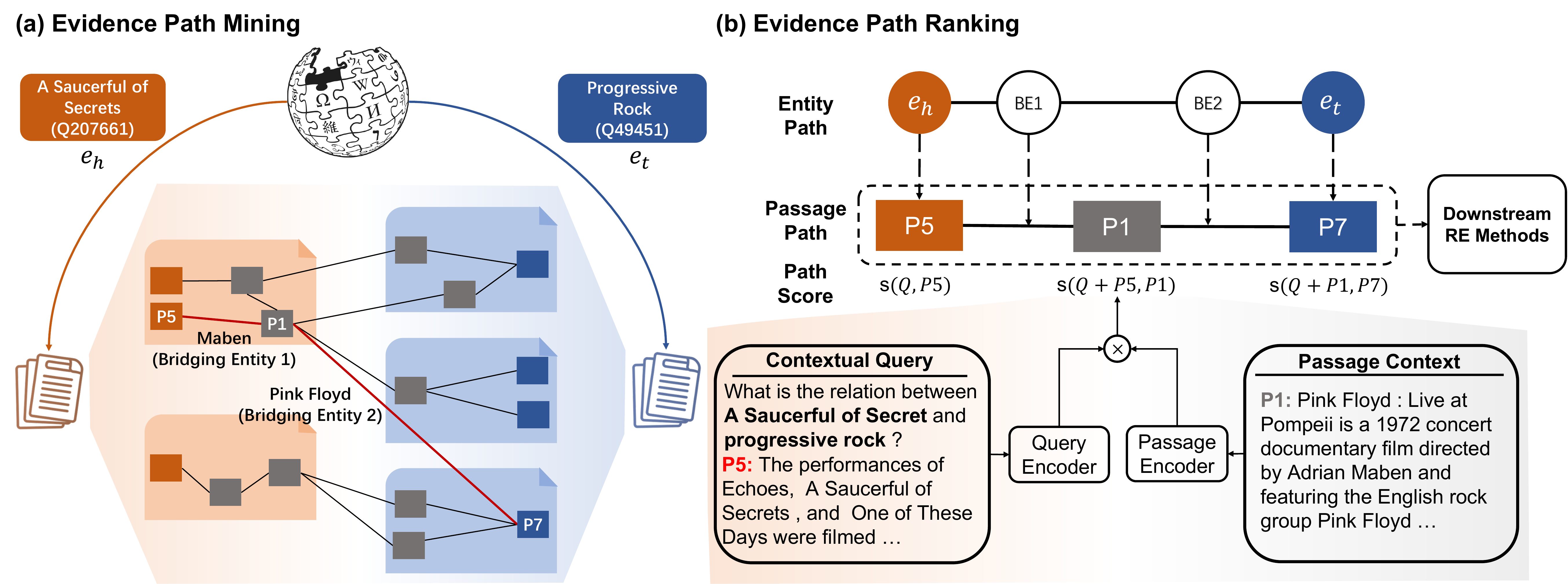}
    \caption{
    Overview of the evidence retrieval method \modelname.
    Given head and tail entities \textit{A Saucerful of Secrets} and \textit{Progressive Rock}, \modelname extracts documents with mentions of them and then builds a multi-document passage graph as shown in subfigure (a).
    A 3-hop candidate evidence path is marked in red, linked by two bridging entities, \textit{Maben} and \textit{Pink Floyd}.
    This candidate evidence path is then scored by a contextual dense retriever, as shown in subfigure (b).
    The sequential scoring process takes a contextual query and the next hop passage as input.
    Evidence paths ranked as top-K will be further adjusted in length and used as evidence for downstream RE methods.
    }
    \label{fig:method}
\end{figure*}

Against these challenges, we propose a dedicated solution \modelname \modelnameexplain, which extracts evidence from a large corpus by multi-hop dense retrieval.
As illustrated in \Cref{fig:method}, \modelname is composed of two steps: evidence path mining and evidence path ranking.
In evidence path mining, we first construct a multi-document passage graph, where passages are linked by edges corresponding to shared entities.
Then, we use a graph traversal algorithm to mine passage paths from head to tail entities.
This step greatly reduces the size of candidate evidential passages.
In evidence path ranking, we rank the mined paths based on their relevance.
We explore different dense retrievers widely used in open-domain question answering (ODQA) as scorers for evidence paths and further propose a contextual dense retriever better suited for multi-hop relation inference.
Finally, the top-K evidence paths are selected as input for downstream relation extraction models.
\modelname is flexible and can be used with any models designed for long-context RE.

The contributions of this work are two-fold.
First, we propose a multi-hop evidence retrieval method for cross-document RE and show that high-quality evidence retrieval benefits end-to-end RE performance.
Second, we explore multiple widely-used retrievers in our setting and further develop a contextual dense retriever for multi-hop reasoning.
Our contributions are verified in both closed and open settings of CodRED \cite{yao-etal-2021-codred}.
We observe that \modelname outperforms other evidence retrieval baselines and boosts end-to-end RE performance with various downstream RE methods.

\section{Related Works}
We discuss two topics of research that are closely relevant to this study.

\xhdr{Relation Extraction}
Recent studies on RE are typically based on datasets with sentence-level~\cite{zhang-etal-2017-position,hendrickx-etal-2010-semeval} or document-level~\cite{yao-etal-2019-docred,walker2005ACE} context.
Except for manually annotated datasets, another part of RE focuses on distantly labeled datasets~\cite{riedel2010modeling}, which takes a bag of sentences mentioning the same entity pair as input~\cite{mintz-etal-2009-distant}. 
However, context lengths in these datasets are considerably smaller than that of cross-document RE, an understudied setting that we focus on.
\citet{yao-etal-2021-codred} first proposed the cross-document RE task and provided a manually constructed dataset CodRED. 
\citet{wang2022entity} proposed entity-centric snippet selection and cross-path relation attention to enhance performance in cross-document RE.
Nevertheless, this work targets the closed setting where evidential context has been given instead of the more challenging and realistic open setting we investigate in this paper.

\xhdr{Evidence retrieval for RE} Evidence retrieval has shown to be effective in document-level RE.
\citet{huang-etal-2021-three} proposed a simple heuristic method to select evidence sentences from documents.
\citet{huang-etal-2021-entity} used evidence as auxiliary supervision to guide the model in finetuning.
\citet{xie2022eider} developed a lightweight model to extract evidence sentences.
However, these works limit evidence retrieval to a single document and are infeasible to scale up to the cross-document setting.
In cross-document RE, \citet{yao-etal-2021-codred} proposes a heuristic simple way to extract evidence, which selects text paths based on the occurrence count of head and tail entities and selects snippets around entity mentions as evidence.
\citet{wang2022entity} enhances evidence selection in the closed setting as entity-centric snippet selection.
However, neither method considers multi-hop reasoning, which is vital for cross-document RE.

\xhdr{Dense Retrieval} 
Dense retrieval is a fast-developing research topic summarized adequately by the latest surveys, \citet{zhao2022dense} and \citet{zhu2021retrieving}.
Therefore, we only provide a highly selected review.
\citet{karpukhin-etal-2020-dense} proposed a dense passage retriever (DPR) with bi-encoder encoding queries and contexts separately.
\citet{lee-etal-2021-phrase} extended DPR to phrase retrieval and conducts passage retrieval based on phrases.
However, these two methods do not directly support multi-hop retrieval.
\citet{xiong2021answering} proposed a multi-hop retriever with a shared encoder and query-augmented methods.
Nevertheless, its experiments are constrained to two-hop reasoning, although it can theoretically be extended to more hops.
Besides, generative retrieval methods can potentially serve as retrievers in our method.
We leave this direction as a feature study.
However, dense retrievers designed for open-domain question answering (ODQA) can not directly apply to evidence retrieval in cross-document RE because the semantics of queries in this setting are much sparser.
In this work, we adapt representative dense retrieval methods and further develop a variant of DPR specifically for evidence retrieval in cross-document RE.

\section{Methods}
In this section, we describe \modelname, a multi-hop evidence retrieval method for cross-document RE.
We will introduce the preliminaries (\Cref{sec:method_def}), proposed evidence retrieval (\Cref{sec:method_path_mining} - \Cref{sec:method_input_prep}), and downstream RE methods we explored (\Cref{sec:method_downstream_re}).

\subsection{Preliminaries}\label{sec:method_def}
\xhdr{Problem Definitions} The input of cross-document RE consists of a head entity $e_h$, a tail entity $e_t$, and a corpus of documents.
The documents are annotated with mentions of entities.
Cross-document RE aims to infer the relation $r$ between the head and tail entities from a candidate relation set $\mathbf{R}$.
Following \citet{yao-etal-2021-codred}, cross-document RE has two settings.
In the \textbf{closed setting}, only the related documents are provided to the models, and the relations are inferred from the provided documents.
While in the more challenging and realistic \textbf{open setting}, the whole corpus of documents is provided, and the model needs to efficiently and precisely retrieve related evidence from the corpus.
We conduct experiments in both settings.

\xhdr{Method Overview}
We divide cross-document RE into two phases: evidence retrieval and relation inference. \textbf{Evidence retrieval} aims to retrieve evidential context that is short enough to meet the input length constraint of downstream RE models while providing sufficient information for relation inference.
\textbf{Relation inference} determines the relations between pairs of entities based on the retrieved evidence.
Our main contribution is a multi-hop evidence retrieval method consisting of a graph-based evidence path mining algorithm and a path ranking method with multi-hop dense retrievers as scorers.
\Cref{fig:method} shows our proposal in detail with the same input example in \Cref{fig:data_case}.
We assume that the evidence is represented as an \textbf{evidence path}, i.e., a chain of passages linked by bridging entities.
This evidence path begins with passages containing head entities (i.e., head passages) and ends with passages containing tail entities (i.e., tail passages).
This assumption is also widely adopted in multi-hop reasoning in ODQA~\cite{feldman-el-yaniv-2019-multi,feng-etal-2022-multi}.
Subfigure (b) in \Cref{fig:method} displays an evidence path with four passages linked by three entities.
We build a multi-document passage graph and run the graph traversal algorithm to find candidate evidence paths, shown in the subfigure (a) in \Cref{fig:method} (\Cref{sec:method_path_mining}).
We rank all candidate evidence paths using multi-hop dense retrievers as scorers, shown in subfigure (b) in \Cref{fig:method} (\Cref{sec:method_path_ranking}).
The top-K evidence paths are selected and further prepared as input of downstream RE models (\Cref{sec:method_input_prep}).
Our evidence retrieval method is agnostic to downstream RE models.
Therefore, we adopt previous RE models for relation inference (\Cref{sec:method_downstream_re}).

\subsection{Evidence Path Mining}\label{sec:method_path_mining}
Evidence path mining aims to efficiently extract multi-hop evidence paths that align with our assumptions from an open set of documents.

For head-to-head comparison, we follow the text path assumption in \citet{yao-etal-2021-codred}, i.e., a candidate evidence path can only go across two documents containing head and tail entities, respectively.
Therefore, we only keep documents containing at least one mention of head and tail entities to build a multi-document passage graph consisting of three types of nodes: head passages, tail passages, and other passages that do not contain head or tail mentions.
If two passages share mentions of one entity, we create an edge marked for this entity between them.
There may be multiple edges between two passages if they share multiple entities.
Given this graph as an input, our algorithm finds all paths from head passages to tail passages as evidence paths with the graph traversal method.

Specifically, we employ depth-first search, an efficient unsupervised path mining algorithm, to find evidence paths on the previously obtained passage graph.
Graph traversal begins at a head passage and ends at a tail passage.
The detailed algorithm is described in \Cref{alg:graph_traversal} in \Cref{sec:appendix_alg}. 
To eliminate repetition and ensure that an evidence path is mined from a text path as the same setting by \citet{yao-etal-2021-codred}, we enforce several additional constraints:
\begin{itemize}[leftmargin=1em]
    \setlength\itemsep{0em}
    \item An evidence path should not contain head or tail entities in the middle of the path.
    \item There should be no repeated passages or repeated bridging entities in the path.
    \item Max lengths of paths should be less than a predefined length $H$.
    \vspace{-0.5em}
\end{itemize}
These constraints encourage our algorithm to prioritize shorter paths.
They also help to improve efficiency and mine more meaningful evidence paths, as two entities are more directly related within a shorter evidence path.
A probing analysis in \Cref{tab:path_mining} shows that most of the evidence can be recalled by paths with less than five hops.
This insight is also exacerbated in other works focusing on multi-hop reasoning~\cite{yang-etal-2018-hotpotqa,xiong2020answering}.

\subsection{Adapting Retrievers for Path Ranking}\label{sec:method_path_ranking}
We adapt dense retrievers to rank evidence paths and develop a contextual variant to overcome sparse query semantics for evidence retrieval.

\subsubsection{Dense passage retriever (DPR)}\label{sec:method_dpr}

DPR is a bi-encoder model first developed as the retrieval component in ODQA.
It uses a dense encoder $E_P(\cdot)$ to map text passages into offline low-dimension vector indices.
During runtime, it uses another query encoder $E_Q(\cdot)$ to retrieve the top-K most similar passages based on maximum inner product search (MIPS):
\begin{equation}\label{eq:inner_product}
    \textrm{sim}(q, p) = E_Q(q)^TE_P(p).
\end{equation}
Two encoders are independent BERT models~\cite{devlin-etal-2019-bert}, and the representations of \texttt{[CLS]} tokens are used to represent the query or passage.

DPR is trained with a contrastive loss.
Let $T=\{\langle q_i, p_i^+,\{p^-_{i,j}\}_{j=1}^{n}\rangle\}_{i=1}^N$ be the training corpus where $q, p^+, p^-,n, N$ are queries, positive, negative passages, number of negative passages and samples, the loss function is formulated as:
\begin{align*}\label{eq:contrastive_loss}
    &l(T_i) =
    &-\log{\frac{e^{\text{sim}(q_i, p^+_i)}}{e^{\text{sim}(q_i, p^+_i)} + \sum_{j=1}^ne^{\text{sim}(q_i, p^-_{i,j})}}}.
\end{align*}
The negative passages can be other passages in the same batch or hard ones mined using BM25.

\subsubsection{Adapting DPR in Cross-document RE}\label{sec:method_adapt_dpr}
We employ a dense retriever to measure similarities between entity pairs and potential evidence passages.
However, although DPR is widely used and proven effective in ODQA, using it directly for cross-document RE has two limitations.
First, queries in ODQA are richer in semantics, while queries in cross-document RE only focus on identifying the relations between head and tail entities.
Accordingly, we use the following simple template to transform entity pairs into semantic queries:
\begin{tcolorbox}
\small
    What is the relation between \textbf{\color{yaleblue}Head Entity} and \textbf{\color{yaleblue}Tail Entity}?
\end{tcolorbox}

Second, DPR does not consider more than one positive passage, while evidence paths in cross-document RE consist of multiple passages.
To address this issue, we extend the training corpus of DPR to multiple positive scenarios, where $T=\{\langle q_i, \{p_{i,k}^+\}_{k=1}^{m},\{p^-_{i,j}\}_{j=1}^{n}\rangle\}_{i=1}^N$ and m denotes to the number of positive evidence, and the loss function is formulated as:
\begin{equation}\label{eq:contrastive_loss_multipos}
\begin{aligned}
    l(T_i) =-\sum_{k=1}^m\log{\frac{e^{\textrm{sim}(q_i, p^+_{i,k})}}{e^{\textrm{sim}(q_i, p^+_{i,k})} + \sum_{j=1}^ne^{\textrm{sim}(q_i, p^-_{i,j})}}}.
\end{aligned}
\end{equation}

As for inference, we use DPR as a scorer to rank evidence paths.
For an evidence path $P=[p_i]_{i=1}^H$ and a query $q(e_h, e_t)$, the ranking score is calculated as the average similarity between the query and all passages:
\begin{equation}\label{eq:pair_scoring}
    s(q(e_h, e_t), P) = \frac{1}{\vert P\vert}\sum_{p_i \in P} \textrm{sim}(q(e_h, e_t), p_i)
\end{equation}

\subsubsection{Contextual DPR}\label{sec:method_cdpr}

Although DPR can be adapted as a scorer for ranking evidence paths in cross-document RE,
it is not specifically designed for multi-hop retrieval.
Besides, queries in our setting are significantly briefer than those in ODQA and may not be sufficient for retrieving relevant passages.
Therefore, we develop a contextual DPR as a multi-hop retriever.

\xhdr{Training}
To enrich the semantics of queries and enable multi-hop reasoning, we augment the original training corpus by data augmentation, where we concatenate the original queries and positive evidence to form new queries.
Specifically, for training data $T=\{\langle q(e_h, e_t), \{p_{i,k}^+\}_{k=1}^m, \{p_{i,j}^-\}_{j=1}^n\rangle\}_{i=1}^N$ where N is the size of data, we augment the query with one of the positive passage.
Therefore, the augmented sample is $T^\prime = T\cup \{\langle q(e_h, e_t)\oplus p_{i,l}, \{p_{i,k}^+\}_{k=1,k\neq l}^m, \{p_{i,j}^-\}_{j=1}^n\rangle\}_{l=1}^m$, where $\oplus$ denotes the string concatenation.
We follow the same negative sampling strategy in the original DPR and train this contextual variant with the same loss function as \Cref{eq:contrastive_loss_multipos}.

\xhdr{Inference} We conduct a sequential scoring process with contextual DPR as the scorer for evidence path ranking.
Denoting $P=[p_i]_{i=1}^H$ as an evidence path consisting of $H$ passages, this scoring function calculates sequential similarities between augmented queries and the next hop of passage:
\begin{equation}\label{eq:sequential_scoring}
    \begin{aligned}
    s(q(e_h, e_t), P) = \frac{1}{\vert P\vert} (\textrm{sim}(q(e_h, e_t), p_1)\\
    +\sum_{i=2}^{\vert P\vert}\textrm{sim}(q(e_h, e_t)\oplus p_{i-1}, p_i)).
    \end{aligned}
\end{equation}
This sequential scoring function \Cref{eq:sequential_scoring} can take advantage of additional context in the query.
Furthermore, embeddings of all augmented queries can be calculated offline to ensure efficiency.

\subsection{Input Preparation}\label{sec:method_input_prep}

To make the evidence path suitable as input for a downstream RE model with maximum input sequence length $L$, we need to transform it into a token sequence that fits within $L$.
If the length of all evidence exceeds $L$, we iteratively drop sentences containing the least number of mentions until the total length fits in $L$ while avoiding dropping sentences containing mentions of head or tail entities.
If the length of all evidence is smaller than $L$, we augment each passage in the evidence path by evenly adding more tokens from the preceding and succeeding snippets until the total length meets $L$, which is the same strategy adopted in \citet{yao-etal-2021-codred}.
After this length adjustment, all passages in the evidence path are concatenated in order as input for downstream RE methods.

\subsection{Downstream RE Methods}\label{sec:method_downstream_re}
Downstream RE methods are the last component in cross-document RE, which takes head and tail entities and evidence context extracted by previous steps as input and conducts relation inference.
As our method focuses on evidence retrieval that is agnostic to the RE methods, we use the same RE methods in the previous cross-document RE benchmarks for head-to-head comparison, which are described in detail in \Cref{sec:experimental_setup}.

\section{Experiments}
This section presents an experimental evaluation of \modelname for evidence retrieval and end-to-end RE performance.
We introduce the experimental setup (\Cref{sec:experimental_setup}), main results (\Cref{sec:results}), and ablation study on incorporated techniques (\Cref{sec:ablation}).

\subsection{Experimental Setup}\label{sec:experimental_setup}

\xhdr{Datasets} We conduct our experiments on the cross-document RE benchmark CodRED~\cite{yao-etal-2021-codred} built from the English Wikipedia and Wikidata.
CodRED contains 5,193,458 passages from 258,079 documents with mention annotations of 11,971 entities.
There are 4,755 positive relational facts annotated for 276 relations and 25,749 NA (Not Available) relational facts.
We experiment in both the closed and open settings discussed in~\Cref{sec:method_def}.
In the closed setting, the context is organized in text paths, i.e., a pair of documents containing head and tail entities.
A relational fact corresponds to multiple text paths.
In the open setting, the context is a subset of Wikipedia documents.
A subset of CodRED also has human-annotated sentence-level evidence annotations, which can be used as fine-tuning data for evidence retrieval.
More detailed statistics can be found in \Cref{tab:statistics}.

\xhdr{Metrics}
We report the F1 score and the area under the precision and recall curve (AUC) as end-task RE metrics which are the same as \citet{yao-etal-2021-codred}.
Scores on the test set are obtained from the official CodaLab competition of CodRED.
We report path- and passage-level recall to evaluate the evidence retrieval component.
The path-level recall is the proportion of evidence paths fully extracted by methods, while passage-level recall only considers the proportion of evidence passages recalled by methods.
The path-level recall is more strict but has closer correlations to downstream RE performance.
To further investigate the different performances of evidence retrieval with short and long paths, we provide recall among evidence paths with $\leq 3$ and $\textgreater 3$ hops, respectively.

\begin{table*}[t]
\centering
\begin{threeparttable}
\setlength{\tabcolsep}{1mm}{
{
\small
\begin{tabular}{p{8mm}llcccc|cccc}
\toprule
\multicolumn{3}{c}{\textbf{Method}}  & \multicolumn{4}{c}{\textbf{Closed Setting}}& \multicolumn{4}{c}{\textbf{Open Setting}} \\
\multicolumn{2}{l}{\textit{Evidence Retriever}} & \textit{RE model}& F1 (D) & AUC (D) & F1 (T) & AUC (T) & F1  (D) & AUC  (D) & F1 (T) & AUC (T) \\
\midrule
/ & & Longformer+ATT & 48.96 & 45.77 & 49.94 & 45.27 & 44.38 & 37.04 & 42.79 & 37.11 \\
BridgingCount & & BERT+CrossATT & 61.12 \tnote{\dag} & 60.91\tnote{\dag} & 62.48\tnote{\dag} & 60.67\tnote{\dag} &  $--$\tnote{\ddag} & $--$\tnote{\ddag} & $--$\tnote{\ddag} & $--$\tnote{\ddag} \\
\midrule
\multicolumn{2}{l}{\multirow{3}{*}{Snippets}} & EntGraph & 30.54\tnote{\dag} & 17.45\tnote{\dag} & 32.29\tnote{\dag} & 18.94\tnote{\dag} & 26.45\tnote{\dag} & 14.07\tnote{\dag} & 28.70\tnote{\dag} & 16.26\tnote{\dag} \\
& & BERT+ATT & 51.26\tnote{\dag} & 47.94\tnote{\dag} & 51.02\tnote{\dag} & 47.46\tnote{\dag} & 47.23\tnote{\dag} & 40.86\tnote{\dag} & 45.06\tnote{\dag} & 39.05\tnote{\dag} \\
& & BERT+CrossATT & 59.40 & 55.95 & 60.92 & 59.47 & 51.69 & 49.59 & 55.90 & 51.15 \\
\midrule
\multicolumn{2}{l}{\multirow{2}{*}{\modelname}} & BERT+ATT & 57.16 & 54.43 &  57.47 & 53.18 & 51.83 & 46.39 & 52.59 & 47.08 \\
& & BERT+CrossATT & 61.20 & 59.22 & 62.53 & 61.68 & 53.06 & 51.00 & 57.88 & 53.30\\
\bottomrule
\end{tabular}
}}
\begin{tablenotes}
    \setlength\itemsep{-3.5pt}
    \raggedright
    \item[\dag] {\footnotesize indicates results collected from \citet{yao-etal-2021-codred} and \citet{wang2022entity}}.
    \item[\ddag] {\footnotesize BridgingCount is designed for the closed setting and is unable to scale up to the open setting}
    \end{tablenotes}
\end{threeparttable}
\caption{
End-to-end RE results of \modelname with contextual DPR and baselines on dev and test sets of CodRED.
We report F1 and AUC in closed and open settings. (D) and (T) refer to the results on dev and test set splits respectively.
Results on the test set are obtained by the official competition of CodRED on the CodaLab.
}
\label{tab:end2end-result}
\end{table*}

\begin{table*}[t]
\centering
\small
\begin{threeparttable}
\setlength{\tabcolsep}{2.5mm}{
\begin{tabular}{llcccccc}
\toprule
&  &\multicolumn{3}{c}{Path-level Recall\dag} & \multicolumn{3}{c}{Passage-level Recall\ddag} \\
& & All & $H_T\textless3$ & $H_T\geq3$ & All & $H_T\textless3$ & $H_T\geq3$ \\
\midrule
\multicolumn{2}{c}{Snippets} & 17.53 & 28.73 & 11.22 & 51.98 & 60.64 & 49.72 \\
\midrule
\cellcolor{blue!10}& All paths & 53.74 & $--$ & $--$ & $--$ & $--$ & $--$ \\
\cellcolor{blue!10}& Random & 14.07 & 22.46 & 9.34 & 44.47 & 50.29 & 42.96 \\
\cellcolor{blue!10}& w/ BM25 & 22.73 & 35.48 & 15.56 & 51.78 & 60.55 & 49.49 \\
\cellcolor{blue!10}& w/ MDR & 22.71 & \textbf{44.52} & 8.18 & 41.57 & 47.10 & 40.12 \\
\cellcolor{blue!10}& w/ DPR & \underline{23.93} & 37.30 & \underline{16.41} & \underline{53.52} & \underline{62.35} & \underline{51.22} \\
\cellcolor{blue!10}\multirow{-6}{*}{\rotatebox[origin=c]{90}{\textit{\modelname ($H$=3)}}} & w/ Contextual DPR & \textbf{25.88} & \underline{39.68} & \textbf{18.10} & \textbf{55.04} & \textbf{65.48} & \textbf{52.31} \\
\midrule
\cellcolor{blue!10}& All paths & 67.79 & $--$ & $--$ & $--$ & $--$ & $--$ \\
\cellcolor{blue!10}& Random & 13.93 & 20.32 & 10.32 & 44.78 & 49.23 & 43.62 \\
\cellcolor{blue!10}& w/ BM25 & 23.64 & 36.27 & 16.54 & 54.25 & 62.00 & 52.22 \\
\cellcolor{blue!10}& w/ MDR & 23.80 & \textbf{45.01} & 9.00 & 43.99 & 48.23 & 41.54 \\
\cellcolor{blue!10}& w/ DPR & \underline{24.85} & 37.86 & \underline{17.52} & \underline{56.02} & \underline{64.27} & \underline{53.87} \\
\cellcolor{blue!10}\multirow{-6}{*}{\rotatebox[origin=c]{90}{\textit{\modelname} ($H$=4)}} & w/ Contextual DPR & \textbf{27.18} & \underline{41.92} & \textbf{18.53} & \textbf{57.12} & \textbf{67.73} & \textbf{54.02} \\
\bottomrule
\end{tabular}
}
\begin{tablenotes}
    \setlength\itemsep{-2pt}
    \raggedright
    \item[\dag] {\footnotesize The proportion of fully extracted evidence paths}.
    
    \item[\ddag] {\footnotesize The proportion of recalled evidence passages}.
    \end{tablenotes}
\end{threeparttable}
\caption{
Evidence retrieval results of \modelname compared with baselines and different variants on the dev set of the CodRED evidence dataset.
$H$ and $H_T$ denote the maximum hop number of path mining and hop number in gold evidence, respectively.
The best scores are identified in \textbf{bold}, and the second best scores are \underline{underlined}.
}
\label{tab:retrieval-result}
\end{table*}

\xhdr{End-to-end Baselines} We compare \modelname with the retrieval baseline \textbf{Snippet} proposed by \citet{yao-etal-2021-codred}.
This method first retrieves text paths based on the rank of counts of head and tail entities, then extracts 256-token snippets around the first head and tail mentions in the text path as evidence for downstream RE methods.
\citet{wang2022entity} proposes a bridging entity-centric method \textbf{BridgingCount}, which first retrieves sentences with a count of bridging entities and then reorders them with SentenceBERT~\cite{reimers-2020-multilingual-sentence-bert}.
However, this method only works for the close setting, so we only compare end-to-end RE performance in the close setting with it.
We then compare end-to-end RE performance of \modelname with the Snippets baseline in both closed and open settings based on three RE methods:\footnote{We rename methods in \citet{yao-etal-2021-codred} and \citet{wang2022entity} to avoid name confusion in this paper.}
(1) \textbf{EntGraph}~\cite{yao-etal-2021-codred} is a graph-based method that first infers intra-document relations and then aggregates them to obtain cross-document relations.
This method is named as Pipeline in \citet{yao-etal-2021-codred}.
(2) \textbf{BERT+ATT}~\cite{yao-etal-2021-codred} uses a BERT encoder and selective attention to encode evidence. This method is named as End2End in \citet{yao-etal-2021-codred}.
(3) \textbf{BERT+CrossATT}~\cite{wang2022entity} enhances \textbf{BERT+ATT} via introducing a cross-path entity relation attention.
(4) \textbf{Longformer+ATT} is a variant that replaces the BERT encoder in the BERT+ATT baseline with Longformer~\cite{beltagy2020longformer} so that it can encode two documents at once without evidence retrieval.

\xhdr{Scorer Ablations} We compare the proposed contextual DPR with four scorers in evidence path ranking:
(1) \textbf{Random} is a baseline randomly selecting top-K evidence paths without ranking.
(2) \textbf{BM25}~\cite{robertson1996okapi} is a widely-used sparse information retrieval function based on the bag-of-words model.
(3) \textbf{DPR}~\cite{karpukhin-etal-2020-dense} is a dense passage retriever that we adapt to evidence retrieval as described in \Cref{sec:method_adapt_dpr}.
(4) \textbf{MDR}~\cite{xiong2021answering} is a multi-hop dense retriever with query augmentation.

\xhdr{Configurations} We initialize dense retrievers with pretrained checkpoints on ODQA tasks and then finetune them on the evidence dataset of CodRED.
We conduct a probing analysis to decide the proper maximum hop number for \Cref{alg:graph_traversal} and found only 1.6\% of the cases required more than four-hop reasoning in the training split of evidence dataset in CodRED.
Similar conclusions about maximum hop number are also found in multi-hop ODQA, where two-hop reasoning is set as default~\cite{yang-etal-2018-hotpotqa}.
Therefore, we conduct experiments on both three- and four-hop evidence retrieval to evaluate the effectiveness of our method since a larger hop number will only marginally improve performance but significantly increases the computational complexity of \Cref{alg:graph_traversal}.
For consistency with \cite{yao-etal-2021-codred}, we select the top 16 evidence paths from all paths given by \Cref{alg:graph_traversal} in the open setting.
We use grid search to find optimal hyperparameters in all experiments.
Detailed implementation configurations are described in \Cref{sec:implementation}.
\subsection{Results}\label{sec:results}

We report end-to-end RE results of \modelname and evidence retrieval performance in this section.

\xhdr{End-to-end RE}
We employ \modelname with contextual DPR as the scorer when conducting end-to-end RE evaluation according to ablation study in \Cref{sec:ablation}.
Comparison results are shown in \Cref{tab:end2end-result}.
We focus on the \emph{open setting} since it is more realistic and challenging.
\modelname significantly benefits all RE models compared with Snippets on the open setting.
For example, \modelname outperforms Snippets by 7.53\% in test F1 and 8.03\% in test AUC when using BERT+ATT as the RE model.
And \modelname with BERT+CrossATT achieves the best scores in test F1 (57.88\%) and test AUC (53.30\%), which improves about 2 percent compared with Snippets.
The results illustrate the necessity of multi-hop evidence retrieval for inducing cross-document relations in the open setting. 

At the same time, \modelname also leads to improvements in the \emph{closed setting}, showing that \modelname helps RE models ping-point relevant evidence in the limited context.
The most notable improvement is witnessed in comparison between \modelname and Snippets with BERT+ATT, where \modelname improves by 6.45\% in test F1 and 5.72\% in test AUC.
BridgingCount is a retrieval baseline specifically designed for the closed setting, which is slightly outperformed by \modelname in the closed setting with the same RE model.
However, \modelname achieves the highest performance in the open setting while BridgingCount cannot scale up to the open setting.
As for baselines, the Longformer+ATT performs worse than BERT+ATT and other language model methods adopting evidence retrievers.
These observations show that even in the closed setting, evidence retrieval benefits by targeting the supporting evidence in the given context.

\xhdr{Evidence Retrieval}
We also analyze the performance of \modelname with 3- and 4-hop path mining and multiple scorer variants in \Cref{tab:retrieval-result}.
Path-level recalls of the 3- and 4-hop evidence path mining are 53.74\% and 67.79\%, which indicates the best recall a 3-hop (or 4-hop) evidence path mining model can get without any path filtering.
The random baselines are outperformed by \modelname with retrievers, demonstrating the effectiveness of path ranking.
DPR outperforms BM25 and MDR in all settings.
Contextual DPR with 4-hop path mining further improves performance by around 2 percent on average compared with the original DPR and surpasses Snippets by 9.65\% and 5.14\%, contributing to multi-hop evidence retrieval by enriching query context.
We also observe recalling evidence paths with more hops is more challenging, while 4-hop path mining consistently improves on longer paths. 
However, path- and passage-level recalls are not always consistent.
For example, MDR performs extraordinarily well in recalling short paths but fails on most of the longer ones.

\begin{table}[t]
    \centering
    \small
    \begin{tabular}{lcc}
    \toprule
    Context & F1 & AUC \\
    \midrule
    Gold evidence & 52.44 & 51.37 \\
    w/ random augmentation & 48.93 & 46.85 \\
    w/ random drop bridging & 48.79 & 46.85 \\
    \midrule
    \modelname evidence & 50.61 & 49.30 \\
    w/ input preparation & 51.01 & 49.52 \\
    w/ random augmentation & 48.19 & 45.50 \\
    w/ random drop bridging & 48.93 & 49.12 \\
    \bottomrule
    \end{tabular}
    \caption{End-to-end RE results on evidence dev set in CodRED with BERT+ATT.
    We consider gold and \modelname evidence as input and develop two variants to show the importance of precise evidence retrieval.
    }
    \label{tab:importance_of_evidence}
\end{table}

\begin{table}[t]
\centering
\begin{threeparttable}
\small
\setlength{\tabcolsep}{1.5mm}{
\begin{tabular}{p{2mm}lcc|cc}
\toprule
\multicolumn{2}{c}{\textbf{Method}}  & \multicolumn{2}{c}{\textbf{Closed}}& \multicolumn{2}{c}{\textbf{Open}} \\
\multicolumn{2}{l}{\textit{Evidence Retriever}} & F1 & AUC & F1 & AUC \\
\midrule
\multicolumn{2}{l}{Snippets} & 47.23\tnote{\dag} & 40.86\tnote{\dag} & 45.06\tnote{\dag} & 39.05\tnote{\dag} \\
\midrule
\cellcolor{blue!10}& Random & 52.12 & 48.34 &  48.92 & 42.13\\
\cellcolor{blue!10}& w/ BM25 & 52.41 & 49.02 &  49.88 & 43.31\\
\cellcolor{blue!10}& w/ MDR&  51.40 & 48.27 & 49.11 & 42.43 \\
\cellcolor{blue!10}& w/ DPR&  55.08 & \underline{51.66} &  51.37 & 46.51 \\
\cellcolor{blue!10}\multirow{-5}{*}{\rotatebox[origin=c]{90}{{\textit{$H$=3}}}} &  w/ Contextual DPR & \textbf{57.65} & 51.08 & \underline{52.32} & \underline{46.56} \\
\midrule
\cellcolor{blue!10}&Random  & 51.78 & 47.09 & 48.22 & 41.65 \\
\cellcolor{blue!10}& w/ BM25 & 52.63 & 49.21 & 50.32 & 44.17 \\
\cellcolor{blue!10}& w/ MDR &  50.98 & 48.01 & 48.97 & 42.59 \\
\cellcolor{blue!10}& w/ DPR &  53.21 & 48.75 &  51.41 & 45.25 \\
\cellcolor{blue!10}\multirow{-5}{*}{\rotatebox[origin=c]{90}{{\textit{$H$=4}}}} & w/ Contextual DPR & \underline{57.47} & \textbf{53.18} & \textbf{52.59} & \textbf{47.08} \\
\bottomrule
\end{tabular}
}
\begin{tablenotes}
    \setlength\itemsep{-3.5pt}
    \raggedright
    \item[\dag] {\footnotesize indicates results collected from \citet{yao-etal-2021-codred}}.
    \end{tablenotes}
\end{threeparttable}
\caption{
End-to-end RE results with \modelname and BERT+ATT based on retriever variants on the test set of CodRED.
We report F1 and AUC in closed and open settings.
The best scores are identified in \textbf{bold}, and the second best scores are \underline{underlined}.
}
\label{tab:end2end-ablation}
\vspace{-1em}
\end{table}
\subsection{Ablation Study}\label{sec:ablation}

We provide the following analyses to further evaluate core components of \modelname.

\xhdr{Importance of Precise Evidence Retrieval} We conduct end-to-end evaluations on evidence dev set with BERT+ATT that uses gold and \modelname evidence as input.
We also build variants by randomly augmenting evidence to 512 tokens or dropping bridging evidence.
\Cref{tab:importance_of_evidence} shows random augmentation and bridging drop decreases end-to-end RE performance with both gold and \modelname evidence as input, which suggests the importance of precise evidence retrieval.
We also found input preparation, a local context augmentation, in \modelname will not damage performance since it helps with recall.

\xhdr{Effectiveness of Evidence Path Mining}
\Cref{tab:path_mining} in \Cref{sec:analysis_path_mining} shows an analysis of evidence path mining with different maximum hop numbers $H$.
First, the number of passage and entity paths mined by \Cref{alg:graph_traversal} increases exponentially when $H$ increases, suggesting an exponential complexity on $H$.
However, a small $H$ will be sufficient since the recall of evidence paths increases marginally as $H \geq 3$.
The failure rates are also less than 7.9\% when $H \geq 3$.
In summary, evidence path mining is effective and efficient under these settings.

\xhdr{Scorers}
\Cref{tab:end2end-ablation} shows an ablation study of scorer variants in both settings on BERT+ATT.
Both 3-hop and 4-hop evidence path mining enhance RE performance even with random selection compared with the Snippets baseline, showing our assumption of evidence paths and path mining can benefit end-to-end RE.
We also witness marginal improvements when $H$ increases.
Contextual DPR with 4-hop evidence path mining achieves the best performance on most metrics.
Comparing \Cref{tab:end2end-ablation} along with \Cref{tab:retrieval-result}, we found evidence retrieval methods with higher recalls tend to perform better in end-to-end RE, which supports the claim that evidence retrieval is crucial for cross-document RE. 

\begin{table*}[t]
    \centering
    \small
    \begin{tabularx}{\linewidth}{l>{\raggedright\arraybackslash}X}
    \toprule
    \textbf{Method} & \textbf{Evidence} \\
    \midrule
    Mr.COD & In the 1934 film, ``[One Night of Love](Head Entity)'', her first film for [Columbia](Bridging Entity), she portrayed a small-town girl who aspires to sing opera. [Harry Cohn](Tail Entity), president and head of [Columbia Pictures](Bridging Entity), took an 18\% ownership in Hanna and Barbera's new company. \\
    \midrule
    Snippets & She was nominated for the Academy Award for Best Actress for her performance in ``[One Night of Love](Head Entity)''.  In 1947, Moore died in a plane crash at the age of 48. [Harry Cohn](Tail Entity), president and head of [Columbia Pictures](Bridging Entity), took an 18\% ownership in Hanna and Barbera's new company. \\
    \midrule
    BridgingEnt & In 1947, Moore died in a plane crash at the age of 48. [Harry Cohn](Tail Entity), president and head of [Columbia Pictures](Bridging Entity), took an 18\% ownership in Hanna and Barbera's new company. She was nominated for the Academy Award for Best Actress for her performance in ``[One Night of Love](Head Entity)''. \\
    \bottomrule
    \end{tabularx}
    \caption{Cases of evidence from different methods.}
    \label{tab:case_study}
    \vspace{-1em}
\end{table*}

\subsection{Case Study}

We provide a case of retrieved evidence to demonstrate the interpretability of \modelname. As the example in \Cref{tab:case_study}, \modelname correctly finds evidence describing the head entity and bridging entity Columbia Pictures. At the same time, Snippets and BridgingCount fail to retrieve evidence that can form a reasoning chain. This example demonstrates that Mr. Cod can perform evidentially supported and precise RE.
We will add a thorough version of this analysis to the final version paper to help readers understand how evidence retrieval works.

\section{Conclusion}
We study efficient and effective ways to extract multi-hop evidence for cross-document RE and propose \modelname.
\modelname extracts evidence paths from an open set of documents and ranks them with adapted dense retrievers as scorers.
To overcome the gap between retrieval in ODQA and evidence retrieval for RE, we develop a contextual DPR that augments sparse queries with passage context.
Extensive experiments show high-quality evidence retrieved by \modelname boosts end-to-end cross-document RE performance.
Future works include extending \modelname to more retrieval methods, such as generative dense retrievers.

\section*{Acknowledgement}

We appreciate the reviewers for their insightful
comments and suggestions.
And we appreciate Yuan Yao and other authors of CodRED dataset~\cite{yao-etal-2021-codred} for updating the dataset and sharing baseline codes.
I-Hung Hsu was supported part by the Intelligence Advanced
Research Projects Activity (IARPA), via Contract
No.\ 2019-19051600007, and a Cisco Research Award.
Wenxuan Zhou and Muhao Chen were supported by the NSF Grant IIS 2105329, the Air Force Research Laboratory under
agreement number FA8750-20-2-10002, an Amazon Research Award and a Cisco Research Award.
Mingyu Derek Ma was supported by the AFOSR MURI grant \#FA9550-22-1-0380, the Defense Advanced Research Project Agency (DARPA) grant \#HR00112290103/HR0011260656, and a Cisco Research Award.
Computing of this work was partly supported by a subaward of NSF Cloudbank 1925001 through UCSD.

\section*{Limitations}

Limitations of this work include that we only investigate \modelname on a set of representative single- and multi-hop dense retrievers.
Recent works have proposed more variants of dense retrievers, such as generative retrievers~\cite{lee2022generative,izacard2020leveraging} and multi-hop retrievers~\cite{das-etal-2019-multi,khattab2021baleen}, that can be adapted to use as alternative scorers in \modelname.
Furthermore, we only conduct experiments on three- and four-hop settings.
Although this choice is reasonable and supported by various works, we admit that more hops could be needed in real-world application scenarios, which is understudied in this paper.

\section*{Ethics Statement}
\modelname is designed for retrieving evidence that supports cross-document RE.
However, the evidence retrieved by \modelname is not always factually correct.
This evidence can only be considered as potential context describing facts between given entities.


\bibliography{anthology,custom}
\bibliographystyle{acl_natbib}

\appendix
\begin{center}
    {
    \Large\textbf{Appendices}
    }
\end{center}

\section{Evidence Path Mining Algorithm}\label{sec:appendix_alg}

{\begin{spacing}{0.8}
\begin{algorithm}[t!]
    \small
    \DontPrintSemicolon
    \KwIn{Head entity $e_h$, Tail entity $e_t$, Maximum hop number $H$, A multi-document passage graph $G(N=(d_i,p_j),E=[(d_i, p_m), e_j, (p_k)])$, A set of passages (nodes) with head mentions $S_{e_h}$, A set of passages (nodes) with tail mentions $S_{e_t}$}
    \KwOut{All evidence paths $\mathbf{P}=\{P_l\}_{l=1}^{n_p}$}

     \tcc{Initilize containers}
    paths = list()\tcp{output}
    path\_entities = list()\tcp{briding entities}
    stack = stack()\tcp{backtracking stack} 
    visited = set()\tcp{visited passage set} 
    seen\_path = dict() \tcp{visited passage with a specific start node}
    \tcc{Initialize searching}
    stack.append($e_h$) \\
    visited.add($e_h$) \\
    seen\_path[$e_h$] = list() \\
    path\_entities.add(None) \\
    \tcc{Search with backtracking}
    \While{length(stack) > 0}{
        start = stack[-1] \\
        neighbor\_nodes = get\_neighbor(G, start)
        \If{start not in seen\_path}{
            seen\_path[start] = list()
        }
        g = 0\\
        \For{passage, entity in neighbor\_nodes}{
            \If{passage not in visited and w not in seen\_path[start]}{
                g = g + 1\\
                stack.append(passage)\\
                visited.add(passage)\\
                seen\_path[start].append(passage)\\
                path\_entities.append(entity)\\
                \If{entity in $S_{e_t}$}{
                    paths.append(stack)\\
                    latest\_pop = stack.pop()\\
                    path\_entities.pop()\\
                    visited.remove(latest\_pop)
                }
                break
            }
        }
        \If{g == 0 or length(stack) > $H$}{
            latest\_pop = stack.pop() \\
            path\_entities.pop() \\
            \If{latest\_pop in seen\_path}{
                del seen\_path[latest\_pop]
            }
            visited.remove(latest\_pop)
        }
    }
    \Return{paths}
    
    \caption{Depth-first search for evidence path mining}
    \label{alg:graph_traversal}
\end{algorithm}
\end{spacing}
}

The core algorithm of proposed evidence path mining is a depth-first search described in \Cref{alg:graph_traversal}.
This is a search algorithm based on  backtracking on the multi-document passage graph.
In this algorithm, path searching begins with a set of nodes, i.e., passages, with head mentions $S_{e_h}$.
These nodes are pushed into a stack and the search algorithm extends this stack via adding neighbors of them into this stack.
The algorithm starts backtracking when the current path finds any passages with tail mentions $S_{e_t}$ or the length of current path meets the maximum hop number $H$.
After the graph traversal, all paths linking $S_{e_h}$ and $S_{e_t}$ with a length less than $H$ will be mined and collected as the candidate evidence paths.
This algorithm may fail in some cases.
For example, there may not exist a path within $H$ hop between $S_{e_h}$ and $S_{e_t}$.
The proportion of failure in the CodRED dataset is analyzed in \Cref{sec:analysis_path_mining}.
In these cases, we introduce a redemption way described in \Cref{sec:redemption} to collect candidate paths.

\section{CodRED Dataset}

We collect the CodRED dataset from its official Github repository\footnote{CodRED Github repository: https://github.com/thunlp/CodRED}, which includes relation triplets, evidence dataset, and documents for closed and open settings.
This repo is licensed under the \textit{MIT} license.
Furthermore, CodRED dataset contains processed Wikidata and Wikipedia.
Wikidata is licensed under the \textit{CC0} license.
The text of Wikipedia is licensed under multiple licenses, such as \textit{CC BY-SA} and \textit{GFDL}.

We process the raw data of CodRED as the recipe described in the official Github repository, which includes transforming raw documents and appending them to a Redis server for downstream RE methods.
The original evidence is annotated at the sentence-level while we transform them into passage-level by simply annotating passages containing evidence sentences as evidence passages.   
And we use the same evaluation metric and obtain evaluation scores of the test set from the official CodaLab competition for CodRED.

\Cref{tab:statistics} shows the statistics of the CodRED dataset.
We use two parts of the dataset, including a full dataset and the subset with evidence.
We use the evidence subset to finetune our retriever models.

\begin{table}[thbp]
\small
    \centering
    \begin{tabular}{lccc}
    \toprule
    Split & Triplets & Text paths \\
    \midrule
    Train & 19,461 & 129,548 \\
    Dev & 5,568 & 40,740 \\
    Test & 5,535 & 40,524 \\
    \midrule
    Train (Evidence) & 3,566 & 12,013 \\
    Dev (Evidence) & 1,093 & 3497 \\
    \bottomrule
    \end{tabular}
    \caption{Statistics of the CodRED dataset}
    \label{tab:statistics}
\end{table}

\section{Analysis of Evidence Path Mining}\label{sec:analysis_path_mining}

\Cref{tab:path_mining} shows evidence path mining analysis results on the CodRED evidence dataset.
This analysis focuses on observing recall, failure rate, lengths, and mining speeds of evidence paths along with the increase of maximum hop number.

\begin{table}[thbp]
\centering
\small
\setlength{\tabcolsep}{2mm}{
\begin{tabular}{c|ccccc}
\toprule
$H$ & Recall & Fail & P. Path & E. Path & Speed \\
\hline
2 & 30.4 & 33.6 & 2.3 & 1.1 & 7358 \\
3 & 53.7 & 7.9 & 32.6 & 9.3 & 2735 \\
4 & 67.8 & 4.7 & 390 & 71 & 284 \\ 
5 & 73.1 & 4.4 & 4309 & 528 & 26 \\ \bottomrule
\end{tabular}}
\caption{Analysis of evidence path mining algorithm (\Cref{alg:graph_traversal}) on maximum hop number on the CodRED evidence dataset.
We report the recall and fail rate (\%) of evidence paths, the average number of passage paths (P. Path) and entity paths (E. Path), and processing speed (iter/s).}
\label{tab:path_mining}
\end{table}

\section{Detailed Implementation}\label{sec:implementation}

We describe the detailed implementation of all components in this work.

\subsection{Retrieval methods}\label{sec:retrieval_implementation}

All supervised retrieval methods are trained on the training split of CodRED evidence dataset and tuned hyperparameters on the dev split.

\xhdr{BM25} We use the Python implementation of BM25 algorithm  \textit{Rank\_BM25}\footnote{Rank-BM25 Github Repository: \url{https://github.com/dorianbrown/rank_bm25}}.
We first remove all characters that are neither English characters and numbers from all passages.
And then we tokenize the passages into bags of words with the \textit{word\_tokenize} function in \textit{NLTK}.
We further preprocess by removing stopwords collected in \textit{gensim} and then stem all words with the \textit{PorterStemmer} in \textit{NLTK}.
The queries are preprocessed in the same way.
We use Okapi BM25 to rank the top-K evidence paths for each query and use it as evidence for the next steps.
All third-party APIs we used in this section are run with default parameters.
The ranking score is calculated as the average BM25 score between the query and each passage in the evidence path.

\xhdr{MDR} We develop our adapted MDR based on the MDR official Github repository.\footnote{MDR Github Repository: \url{https://github.com/facebookresearch/multihop_dense_retrieval}}
We train the MDR from the public checkpoint shared in the repository, which is trained from roberta-base.
We use the same shared encoder setting as the original MDR.
The batch sizes for training and inference are 16; the learning rate is 2e-5; the maximum lengths of queries, contexts, and augmented queries are 70,300,350, respectively; The warm-up rate is $0.1$.
We train this model for 5 epochs on 3 NVIDIA RTX A5000 GPUs for 8 hours.
Then we encode all passages with 8 NVIDIA RTX A5000 GPUs.
The queries are first generated with augmented passages and encoded offline before evidence retrieval.

\xhdr{DPR}  We develop our adapted DPR based on the DPR official Github repository.\footnote{DPR Github Repository: \url{https://github.com/facebookresearch/DPR}} We train the DPR from the public checkpoint trained on the NQ dataset with 
 the single adversarial hard negative strategy.
 We follow the configuration \textit{biencoder\_local} in the original DPR training configurations: batch size is 4; learning rate is 2e-5; warmup steps are 1237; number of training epochs is 50; maximum gradient norm is 2.0.
 We run the training on 8 NVIDIA RTX A5000 GPUs for 12 hours.

\xhdr{Contextual DPR} We use the same setting of DPR to train the contextual DPR, except we prolong the training epochs to 70 and reduce the learning rate to 1e-5.
We run the training on 8 NVIDIA RTX A5000 GPUs for 12 hours.

\subsection{Downstream RE methods}\label{sec:downstream_RE_implementation}

We train downstream RE methods on the training split of CodRED and tune hyperparameters on the dev split.
We strictly follow the CodRED recipe so we also introduce path-level and intra-document supervision when training following RE models.

\xhdr{BERT+ATT} We use the same code collected from the official repository\footnote{BERT+ATT Github Repository:\url{https://github.com/thunlp/CodRED}}. We train BERT+ATT from the sketch on 8 NVIDIA RTX A5000 GPUs for 10 hours.
The training and inference batch size is 1; the learning rate is 3e-5; the number of training epochs is 8; The base model is \textit{bert-base-cased} from Huggingface Transformers.

\xhdr{BERT+CrossATT} We use the same code collected from the official repository\footnote{BERT+CrossATT Github Repository:\url{https://github.com/MakiseKuurisu/ecrim}}.
We train BERT+CrossATT from the sketch on 4 NVIDIA RTX A5000 GPUs for 20 hours.
The training and inference batch size is 1; the learning rate is 3e-5; the number of training epochs is 10; The base model is \textit{bert-base-cased} from Huggingface Transformers.

\subsection{Simple redemption of \modelname}\label{sec:redemption}
We assume we can find a chain of passages linked by bridging entities.
However, this assumption does not always hold for the fixed maximum number of hops.
As \Cref{tab:path_mining} shown, the fail rate increases when the maximum hop number decreases.
We use a simple redemption that selects passages containing head and tail mentions as an evidence path when evidence path mining fails.
The fail rate is considerably low ($\leq 5\%$) when the hop number is more than 3.
Therefore, this simple redemption will not affect prove the effectiveness of the evidence path mining algorithm.
Besides, a small portion of entities may appear in many documents, leading to low efficiency of \Cref{alg:graph_traversal}.
Therefore, we propose another simple redemption that selects top 50 documents based on entity count when an entity appears in more than 50 documents.
The text-path selection in \cite{yao-etal-2021-codred} inspires this redemption.

\end{document}